\documentclass{llncs}
\usepackage{llncsdoc}
\usepackage{graphicx}
\usepackage{array}
\usepackage{cite}
\usepackage{hyperref}
\usepackage{subfigure}
\usepackage[T1]{fontenc}
\usepackage[utf8]{inputenc}     
\usepackage[english]{babel}   
\usepackage{float}
\usepackage{multirow}
\usepackage{url}

\newcolumntype{P}[1]{>{\centering\arraybackslash}p{#1}}

\begin{document}
\title{Broad Neural Network for Change Detection in Aerial Images}
\author{Shailesh Shrivastava\inst{1}, Alakh Aggarwal\inst{2} and Pratik Chattopadhyay\inst{2}}
\institute{Indian Institute of Technology, Patna
\and Indian Institute of Technology (Banaras Hindu University), Varanasi}
\maketitle

\section{Abstract}

A change detection system takes as input two images of a region captured at two different times, and predicts which pixels in the region have undergone change over the time period. Since pixel-based analysis can be erroneous due to noise, illumination difference and other factors, contextual information is usually used to determine the class of a pixel (changed or not). This contextual information is taken into account by considering a pixel of the difference image along with its neighborhood. With the help of ground truth information, the labeled patterns are generated. Finally, Broad Learning classifier is used to get  prediction about the class of each pixel. Results show that Broad Learning can classify the data set with a significantly higher F-Score than that of Multilayer Perceptron. Performance comparison has also been made with other popular classifiers, namely Multilayer Perceptron and Random Forest.

\section{Introduction}

Change detection in images has become one of the important aspects in recent years, with a wide range of applications. Various aspects include detection of changes in motion, change in vegetation cover over an area, extent of damage caused due to floods and other natural calamities, etc. The basic objective of change detection is to identify dissimilar regions by comparing a pair of the images of the same area at different points of time.

Over the past few years, a lot of research work has been carried out in this area, and several 
supervised, unsupervised and semi-supervised learning techniques have been developed, each with some pros and cons. While unsupervised learning does not need any training information, supervised learning techniques require substantial amount of ground truth knowledge, 
which may not be conveniently  available always. 
On the other hand, the semi-supervised technique is efficient in a way that it requires only a few labeled data for training. The few labeled patterns of the changed class can help in training a classifier by means of self-training. 

Recently Broad Learning has been proposed in \cite{bls} as an alternative to Deep Learning. It has been seen that, Broad Learning is immensely successful in classifying image feature sets, and in some cases it performs even better than that of Deep Learning. Moreover, training using Broad Learning is significantly faster than that of Deep Learning. In this work, we use a variation of the Broad Learning algorithm proposed in \cite{bls}, 
and compare its results with the Multilayer Perceptron and Random Forest classifiers. 

\section{Literature Survey}

A change detection method essentially does comparison between two images to differentiate between changed and unchanged pixels. Early methods 
are mostly threshold-based in which a difference image $D$ is first computed from the two input images 
say, $X$ and $Y$, 
and a threshold is applied on the difference image to detect the changes. The threshold is chosen empirically as explained by Rosin \cite{r1},\cite{r2}. This is referred to as`simple differencing' \cite{r7}. Two different techniques are usually employed to generate the difference image: Change Vector Analysis (CVA) \cite{r3},\cite{r4},\cite{r5} and Image ratioing \cite{r6}. While the former uses the modulus of the difference between the feature vectors of the each pixel, the latter uses the ratio of the pixel intensities to generate the difference image. However, manual selection of threshold 
is erroneous and is not robust in presence of changes in illumination, and noise.

In view of the two hypothesis used in change detection algorithms, i.e., the null hypothesis $H_0$ corresponding to a no-change decision and the alternative hypothesis $H_{1}$ corresponding to a change decision at a given pixel, the hypothesis to be chosen to best describe the extent of change at a given pixel (say, \emph{n}) is based upon the conditional joint probability density functions $p(X(n),Y(n)|H\textsubscript{0})$ and $p(X(n),Y(n)|H\textsubscript{1})$, based upon the classical framework of hypothesis testing \cite{r8},\cite{r9}.

Later work on change detection include \cite{r10} in which pixels are softly classified into mixture components for different change models by employing the Expectation Maximization algorithm, 
The algorithm computes the optical flow field between a pair of images. This approach seems to be an improvement over the previous techniques as it estimates mis-alignment, as well as the illumination variation between the input images. 
Another change detection algorithm  based on the concept of 'self-consistency' among the regions of an image is proposed in \cite{r11}. This approach uses the Minimum Description Length (MDL) model \cite{r12} for 
selecting the hypothesis that best describes the image pair.

More sophisticated change detection algorithms make use of the spatial and temporal relation between a pixel and its neighboring pixels. The spatial models consider a polynomial function of a pixel coordinate $n$ upon which the intensity values of a block of pixels is fitted. Constant, linear and quadratic models are used on these blocks of pixels via generalized likelihood ratio tests \cite{r13},\cite{r14}. The threshold value can be obtained by t-test for constant models and F-test for linear or quadratic models. Out of the three models, the quadratic model is found to be a better option compared to the constant and linear models due to its higher confidence in prediction. 

Numerous temporal models have been proposed for detecting changes in sequence of images. Variation of pixel intensities over time is modeled as an auto-regressive process \cite{r15}. Mean, variance and correlation coefficient were estimated and used in likelihood ratio tests where the null hypothesis depicts that the image intensities are dependent and the alternative hypothesis depicts them as independent \cite{r16}. 

Another algorithm known as the Wallflower algorithm uses a Wiener filter for predicting a pixel's value from it previous values \cite{r17}. Pixels are classified as changed if the prediction error is much more than the expected error. This algorithm can also be considered as a background estimation algorithm. Due to low accuracy of the linear models, 
nonlinear models are proposed to study the relationship between images \cite{r18}. The optimal nonlinear function is the conditional expected intensity value $X(n)$ given $Y(n)$. An unsupervised approach to estimate the parameters of the nonlinear predictor is proposed using adaptive neural networks \cite{r19}. Those pixels on which the predictor performed poorly are classified as changed. 

The shading models are used in various change detection techniques to produce illumination-invariant algorithms. These algorithms computed ratio of the intensity of corresponding pixels in the two images as given by:
\begin{equation}
    R(n) = \frac{X(n)}{Y(n)}
\end{equation}
and compare it to an empirically determined threshold. Instead of directly comparing the ratio $R(n)$, a new form of the difference image was introduced in \cite{r20} : 
\begin{equation}
    D(n) = \frac{1}{N}\sum_{l\in \Omega\textsubscript{n}} {(R(n) - \mu\textsubscript{n})}^2,
\end{equation}
where $N$ is the number of pixels in a block of pixels $\Omega_n$, $l$ is a pixel in the pixel block $\Omega_n$, and $\mu_n$ 
is given by Equation \ref{mu}:
\begin{equation}\label{mu}
    \mu\textsubscript{n} = \frac{1}{N}\sum_{l\in \Omega\textsubscript{n}} {R(n)}.
\end{equation}
If $D(n)$ is greater than the threshold, it is considered as a changed pixel.

A change detection algorithm based on the shading model is introduced in \cite{r21} that uses the weighted combinations of difference in intensities and textures and the intensity ratio. This algorithm assumes that the texture information is less sensitive to variations in illuminations as compared to intensity difference. Instead of using the intensity ratios, the technique in \cite{r22} 
compares the circular shift moments to detect the changes in image. The reflectance component of the intensity is represented by these moments and are claimed to capture the details of an image more precisely. 

An algorithm using semi-supervised Multilayer Perceptron was employed in \cite{r23} for detecting changes in remotely sensed images. 
Here, the difference image is generated by the CVA technique using the difference operator. The input pattern for each pixel is generated using the pixel intensity of the difference image as well as those within a spatial neighbourhood. Thus the input vectors contain nine components with gray values of a pixel and gray values of its eight neighbouring pixels.  A few initial labeled patterns were identified automatically using the K-means clustering algorithm. A neural network is trained using these labeled patterns initially. Since the input patterns have nine pixel values, so the input layer has nine units and 1 bias unit. The output layer has two units for the unchanged and changed classes. The remaining unlabeled patterns are processed by the perceptron in a semi-supervised manner by obtain a soft class label for each unlabeled pattern in every iteration, and next using these soft labels along with the ground truth to train the network repeatedly till optimal condition is reached. This approach is highly time-intensive, and suffers from the disadvantage that incorrect prediction at an initial iteration, propagates to the successive iterations as well. 

Another change detection technique for remotely sensed images is proposed in \cite{r24} that carries out fusion of spectral and statistical indices in an unsupervised manner. The spectral changes are taken into account via the CVA technique, while the statistical changes are identified using a similarity index based on Kullback Leibler distance. The spectral change information and similarity index are fused using wavelets. A neuro-fuzzy classifier next classifies the fused image into changed and unchanged classes, overcoming mixed pixel problems of the satellite images.

In this work, we aim to study the performance of Broad Learning in classifying the difference image. It has been shown in previous studies \cite{bls} that Broad Learning is faster and also has a similar (sometimes better) level of accuracy as that of Deep Learning in classifying image feature sets. Moreover, Broad Learning algorithm is incremental, i.e., with the availability of more ground truth data, training of the classifier can be accomplished very fast.

\section{Proposed Approach}
\begin{figure}[ht]
    \centering
    \hspace*{-1cm}
    \includegraphics[scale = 0.7]{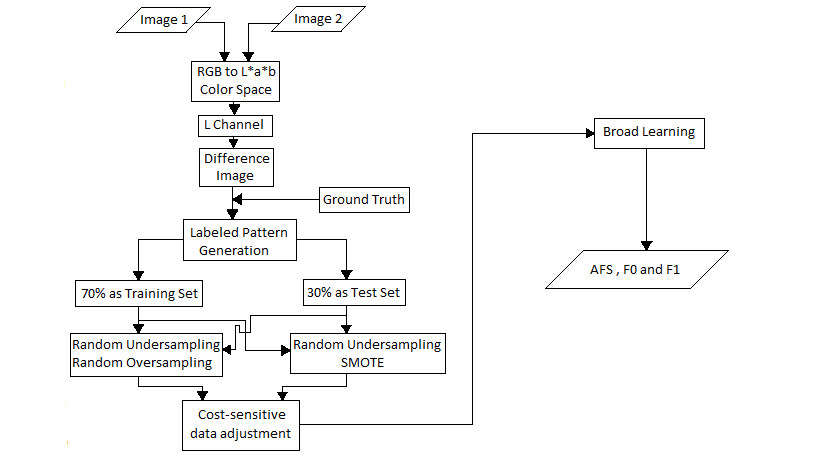}
    \caption{System overview of change detection from a pair of aerial images captured at different time instants}
    \label{flowchart}
\end{figure}

A block diagram of the proposed approach is explained using the block diagram shown in Fig. \ref{flowchart}.

\subsection{Input Pattern Generation}
Consider two co-registered and geometrically corrected RGB images of the same place taken at two different times. Since L*a*b* color space is known to be illumination invariant, the two images are initially transformed from RGB color space to L*a*b* color space. 
For generating the input pattern, a pixel is chosen along with its eight neighboring pixels. 
Thus, the input feature for each pixel is a vector in nine dimensional feature space. The difference image is generated via absolute simple difference of the two images for each channel. 

A general characteristic of change detection data set is that, the ground truth consists of large samples from  unchanged class and very few samples of changed class.  Training any learning algorithm with this data set will cause a high bias for the unchanged class. To account for this problem, we study two different cost-sensitive data adjustment techniques. First the unchanged class of the training is down-sampled using random down-sampling. Next the changed class is up-sampled using a random up-sampling technique, namely,  $Synthetic$ $Minority$ $Over$ $Sampling$ $Technique$ ($SMOTE$). 

We construct various data sets with different $Imbalance$ $Ratios$ (i.e., ratio of the number of majority to the number of minority samples) and study the performance of the classifier for each such case.

\subsection{Classification using Broad Learning}

Broad Learning network \cite{bls} is a recently proposed two-layer deep neural network which consists of input feature map nodes in the input layer, and enhancement nodes in the hidden layer. The original version of Broad Learning has been seen to perform with significant accuracy in classifying some image data sets. Moreover, training using Broad Learning is very fast. Detailed architecture of the network, along with its working principle is given in \cite{bls}. One drawback of the approach in \cite{bls} is that, it does not specify the number of enhancement nodes to be added for optimum performance. This is important since we have limited memory space, and adding enhancement nodes without bounds would cause the memory to overflow.

\begin{figure}
    \centering
    \includegraphics[scale=0.3]{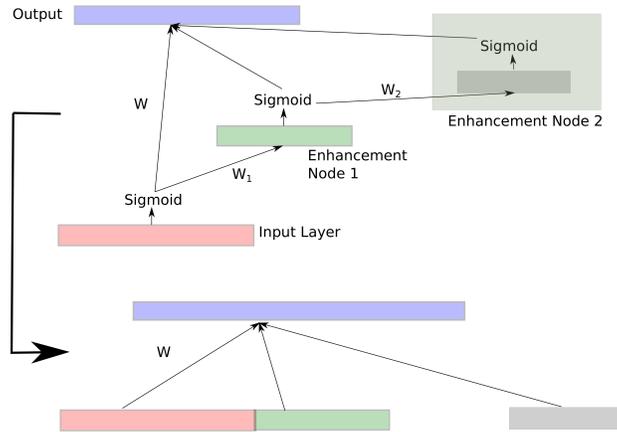}
    \caption{Broad Learning Architecture}
    \label{fig:mod_BL}
\end{figure}

So, we propose a modification to the existing architecture of the Broad Learning System. We continue to recursively add new layer of enhancement nodes till a desired optimum condition is reached. Sparse Autoencoder is employed to find new layer of enhancement nodes at each recursion level with a pre-specified compression rate. 
Finally, all the generated layers are concatenated with the input layer.
The weights from the set of input and enhancement layers to the output layer is next determined using Moore-Penrose Psuedo-inverse, similar to that in \cite{bls}. The scheme is explained with the help of Fig. \ref{fig:mod_BL}.

At each step after adding a new layer of enhancement nodes, we compute the cross-validated average F-Score on classifying the training set. Addition of enhancement nodes is terminated once the average F-Score values in two successive iteration is less than a small threshold value. At this point, further addition of new enhancement nodes does not improve classification accuracy on training set any further, but the network tends to get over-fitted.

\section{Results}

Evaluation of our algorithm is done on a publicly available data set for change detection \footnote[1]{\url{{https://computervisiononline.com/dataset/1105138664}}} which consists of 1000 pairs of 800$\times$600 images, each pair consisting of one reference image and one test image, and the 1000 corresponding 800$\times$600 ground truth masks. The images were rendered using the realistic rendering engine of the serious game Virtual Battle Station 2, developed by Bohemia Interactive Simulations. The data set consists of 100 different scenes containing several objects (trees, buildings, etc.) and moderate  ground relief. Using the reference and test images, the difference image is created and the ground truth masks are used for labeling the data set, and final evaluation of the classifier. To construct the training and test sets, from the ground truth, we randomly select 70\% samples from each class as the training set and remaining 30\% for test set.

We study the effect of various sampling techniques as described above, as well as the effect of decreasing imbalance ratio on the classification performance of Broad Learning. 
Table \ref{BL} shows the F-scores for the two classes along with the Average F Score using the proposed Broad Learning neural network. 
\begin{table}[ht]
\scriptsize
    \caption{Table showing the F-scores for the two classes along with the Average F Score using the Broad Learning Network}
    \label{BL}
    \hspace*{-1cm}
    \begin{tabular}{|P{1.6cm}|P{1cm}|P{2cm}|P{1cm}|P{1cm}|P{1cm}|P{1cm}|P{2cm}|P{1cm}|P{1cm}|P{1cm}|}
    \hline
    \multicolumn{11}{|c|}{Broad Learning approach}\\
        
        \hline
        Imbalance Ratio &
        \multicolumn{5}{p{4cm}|}{Random Undersampling Random Oversampling} & \multicolumn{5}{p{4cm}|}{Random Undersampling SMOTE} \\
        \hline
    
        IR &
        Layers & Compression & AFS & F0 & F1 & Layers & Compression & AFS & F0 & F1 \\
        \hline
        
    \multirow{2}{2em}{1:1} &
    3 & 0.9 & 75.8 & 80 & 72 & 3 & 0.9 & 81.6 & 81.6 & 81.6 \\
    & 3 & 0.7 & 70 & 73 & 67 & 3 & 0.7 & 81.5 & 81.4 & 81.6 \\
    & 5 & 0.9 & 76 & 80 & 72 & 5 & 0.9 & 81.6 & 81.6 & 81.6 \\
    & 5 & 0.7 & 70 & 73 & 67 & 5 & 0.7 & 81.5 & 81.4 & 81.6 \\
    \hline

    \multirow{2}{2em}{2:1} & 
    3 & 0.9 & 79 & 88 & 63 & 3 & 0.9 & 79 & 85 & 66 \\
    & 3 & 0.7 & 74 & 88 & 55 & 3 & 0.7 & 77 & 84 & 63 \\
    & 5 & 0.9 & 79 & 88 & 63 & 5 & 0.9 & 79 & 85 & 66 \\
    & 5 & 0.7 & 74 & 88 & 55 & 5 & 0.7 & 77 & 84 & 63 \\
    \hline
    
    \multirow{2}{2em}{10:1} & 
    3 & 0.9 & 90 & 96 & 27 & 3 & 0.9 & 88 & 95 & 10 \\
    & 3 & 0.7 & 88 & 95 & 10 & 3 & 0.7 & 87 & 95 & 4 \\
    & 5 & 0.9 & 90 & 96 & 27 & 5 & 0.9 & 88 & 95 & 10 \\
    & 5 & 0.7 & 88 & 95 & 10 & 5 & 0.7 & 87 & 95 & 4 \\
    \hline
    
    \multirow{2}{2em}{50:1} & 
    3 & 0.9 & 97 & 99 & 8 & 3 & 0.9 & 0 & 99 & 0 \\
    & 3 & 0.7 & 0 & 99 & 0 & 3 & 0.9 & 0 & 99 & 0 \\
    & 5 & 0.9 & 97 & 99 & 8 & 5 & 0.9 & 0 & 99 & 0 \\
    & 5 & 0.7 & 0 & 99 & 0 & 5 & 0.7 & 0 & 99 & 0 \\
    \hline
    
    \multirow{2}{2em}{100:1} & 
    3 & 0.9 & 99 & 99 & 1 & 3 & 0.9 & 0 & 99 & 0 \\
    & 3 & 0.7 & 0 & 100 & 0 & 3 & 0.7 & 0 & 100 & 0 \\
    & 5 & 0.9 & 99 & 99 & 1 & 5 & 0.9 & 0 & 99 & 0 \\
    & 5 & 0.7 & 0 & 100 & 0 & 5 & 0.7 & 0 & 100 & 0 \\
    \hline
    
    \multirow{2}{2em}{250:1} & 
    3 & 0.9 & 0 & 100 & 0 & 3 & 0.9 & 0 & 100 & 0 \\
    & 3 & 0.7 & 0 & 100 & 0 & 3 & 0.7 & 0 & 100 & 0 \\
    & 5 & 0.9 & 0 & 100 & 0 & 5 & 0.9 & 0 & 100 & 0 \\
    & 5 & 0.7 & 0 & 100 & 0 & 5 & 0.7 & 0 & 100 & 0 \\
    \hline
    
    \end{tabular}
\end{table}

In the table, AFS represents the Average F Score, F0 is the F Score of class 0 and F1 is the F score of class 1.
We observe the effect of varying the number of layers of Enhancement nodes as well as the compression rate while generating a new layer of enhancement node. The set of enhancement nodes generated at each recursion level is termed as a layer, while compression is the ratio of number of nodes in the newly generated layer to that of the previous layer.

\begin{figure}[ht]
\centering
\begin{subfigure}[]{
\centering
    \includegraphics[scale = 0.18]{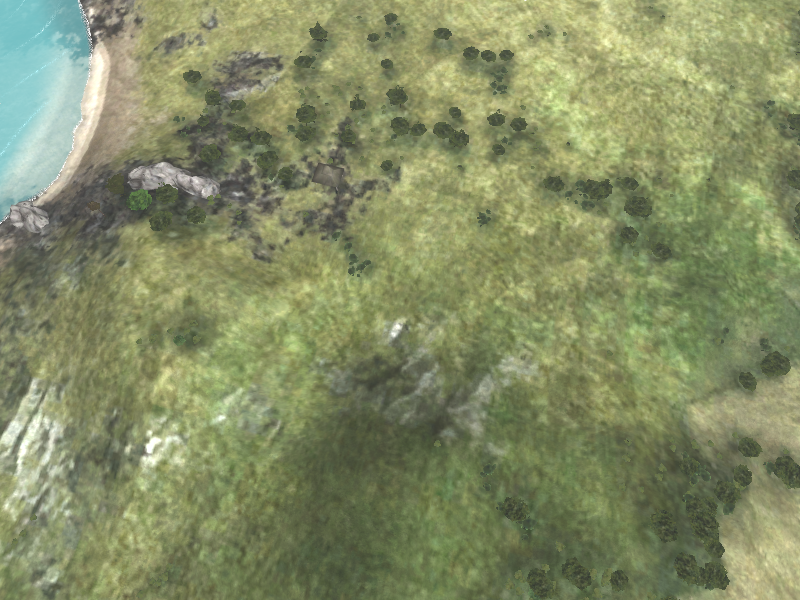}}
\end{subfigure}
\begin{subfigure}[]{
\centering
    \includegraphics[scale = 0.18]{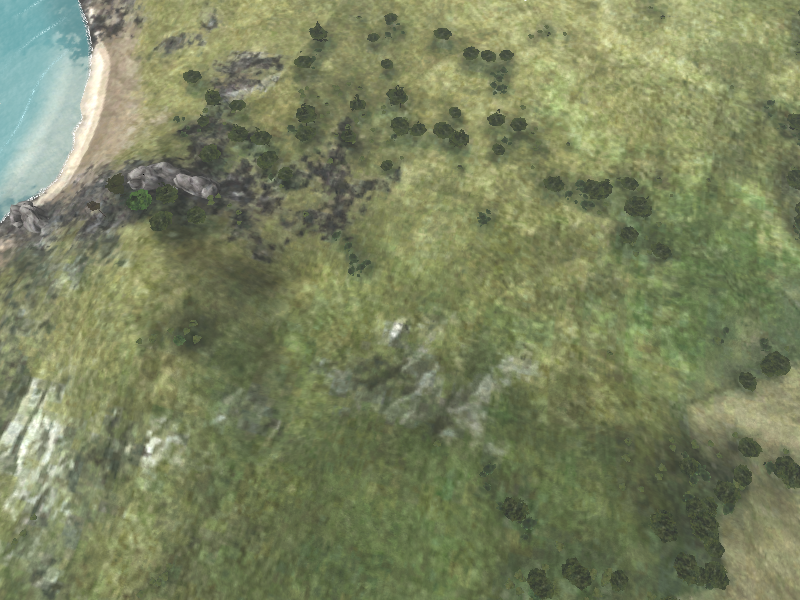}}
\end{subfigure}\\
\begin{subfigure}[]{
\centering
    \includegraphics[scale = 0.18]{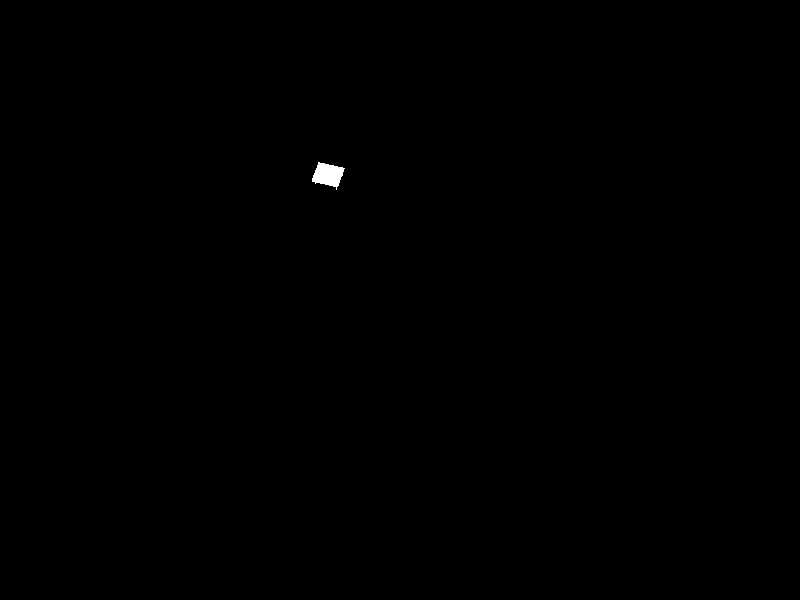}}
\end{subfigure}
\begin{subfigure}[]{
\centering
    \includegraphics[scale = 0.18]{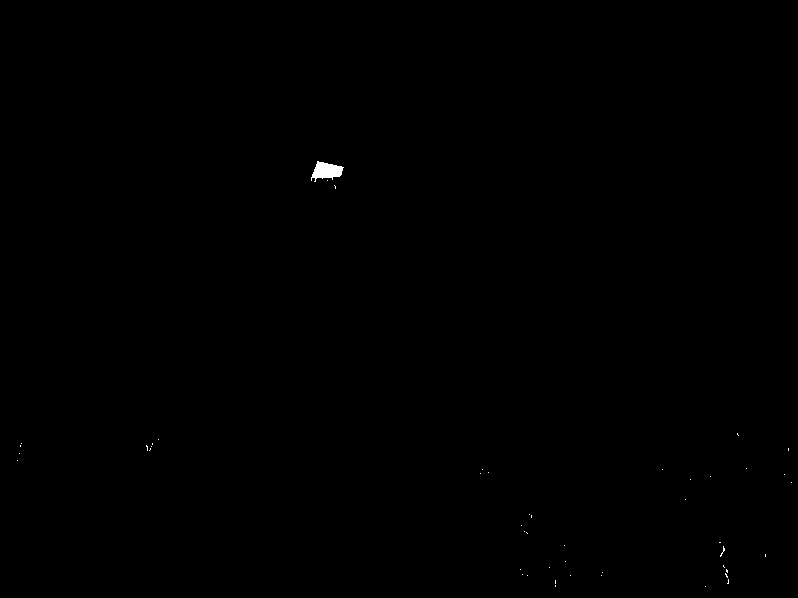}}
\end{subfigure}
\caption{(a) and (b) the two input images, (c) ground truth, (d) predicted changed map from 10:1 imbalanced ratio} \label{fig:op}
\end{figure}

From the table, we conclude that 
SMOTE gives better results and is more effective than random over-sampling. Also, as expected, F-score for minority class (class 1) tends to zero as the imbalance ratio in data set increases. It is seen that, the Broad learning classifier is able to perform reasonably well 
even with the imbalance ratio of 10:1. Also, a higher compression results in more number of nodes in the enhancement layer, and hence, fits the data well. However, the effect of varying the number of layers is not well observed from the data set used in the study. Fig \ref{fig:op} shows the changed map obtained with two pairs of input images using the Broad Learning classifier with 10:1 imbalance ratio.
The first two images shown in Figures \ref{fig:op}(a) and (b) are aerial images of the same place captured at different times, as obtained from the database. Figure \ref{fig:op}(c) shows the ground truth, while Figure \ref{fig:op}(d) shows the change detection map obtained by using the proposed method. Visual observation shows that most of the changed regions are correctly identified by our algorithm, although there exists few false positive cases possibly occurring due to noise/variation of illumination conditions.

We also compare the performance of our approach with two other popularly used classifiers, namely, Random Forest and Multilayer Perceptron. Table \ref{RF} shows the F scores for the two classes along with the Average F Score for both the class using the Random Forest Classifier. 
\begin{table}[ht]
\scriptsize
    \caption{Table showing the F scores for the two classes along with the Average F-Score using the Random Forest Classifier}\label{RF}
    \begin{tabular}{|P{1.6cm}|P{1.1cm}|P{1.1cm}|P{1.1cm}|P{1.1cm}|P{1.1cm}|P{1.1cm}|P{1.1cm}|P{1.1cm}|}
        \hline
        \multicolumn{9}{|c|}{Random Forest Classifier}\\
        
        \hline
        Imbalance Ratio &
        \multicolumn{4}{p{4.35cm}|}{Random Undersampling Random Oversampling} & \multicolumn{4}{p{4.35cm}|}{Random Undersampling SMOTE} \\
        \hline
    
        IR & Trees & AFS & F0 & F1 & Trees & AFS & F0 & F1 \\
        \hline
        
        \multirow{2}{2em}{1:1} & 5 & 43.39 & 60.88 & 17.89 & 5 & 89.29 & 89.99 & 88.59 \\
        & 10 & 37.69 & 67.58 & 7.81 & 10 & 89.40 & 90.17 & 88.63 \\
            
        \hline
    
        \multirow{2}{2em}{2:1} &
            5 & 52.42 & 82.07 & 22.77 & 5 & 89.21 & 93.38 & 85.03 \\
            & 10 & 46.80 & 81.07 & 12.53 & 10 & 88.07 & 92.88 & 83.26 \\
            
        \hline
        
        \multirow{2}{2em}{10:1} & 
            5 & 61.76 & 95.92 & 27.60 & 5 & 82.55 & 97.35 & 67.75 \\
            & 10 & 56.99 & 95.68 & 18.31 & 10 & 80.69 & 97.21 & 64.17 \\
            
        \hline
        
        \multirow{2}{2em}{50:1} &
            5 & 64.66 & 99.14 & 30.18 & 5 & 68.98 & 99.12 & 38.84 \\
            & 10 & 61.99 & 99.12 & 24.85 & 10 & 66.24 &99.12 & 33.36 \\
            
        \hline
        
        \multirow{2}{2em}{100:1} &
            5 & 62.56 & 99.53 & 25.59 & 5 & 63.77 & 99.48 & 28.06 \\
            & 10 & 60.08 & 99.54 & 20.62 & 10 & 59.48 & 99.50 & 19.46 \\
            
        \hline
        
        \multirow{2}{2em}{250:1} &
            5 & 58.98 & 99.78 & 18.18 & 5 & 56.19 & 99.78 & 12.60 \\
            & 10 & 56.51 & 99.80 & 13.22 & 10 & 51.71 & 99.79 & 3.64 \\
            
        \hline
        
    \end{tabular}
\end{table}
Table \ref{MLP} shows the corresponding results for Multilayer Perceptron classifier. 
\begin{table}
\scriptsize
    \caption{Table showing the F scores for the two classes along with the Average F-Score using MLP}\label{MLP}
    \begin{tabular}{|P{1.6cm}|P{1.1cm}|P{1.1cm}|P{1.1cm}|P{1.1cm}|P{1.1cm}|P{1.1cm}|P{1.1cm}|P{1.1cm}|}
    \hline
    \multicolumn{9}{|c|}{Multi Layer Perceptron}\\
        
        \hline
        Imbalance Ratio &
        \multicolumn{4}{p{4.35cm}|}{Random Undersampling Random Oversampling} & \multicolumn{4}{p{4.35cm}|}{Random Undersampling SMOTE} \\
        \hline
    
        IR &
        Alpha & AFS & F0 & F1 & Alpha & AFS & F0 & F1 \\
        \hline
        
    \multirow{2}{2em}{1:1} &
    0.5 & 0 & 66.67 & 0 & 0.5 & 33.34 & 66.65 & 0.02 \\
    & 1 & 36.26 & 5.24 & 67.27 & 1 & 41.48 & 14.51 & 68.45 \\
    \hline

    \multirow{2}{2em}{2:1} & 
    0.5 & 25.03 & 0.06 & 50.01 & 0.5 & 0 & 80.00 & 0 \\
    & 1 & 39.01 & 24.85 & 53.18 & 1 & 0 & 80.00 & 0 \\
    \hline
    
    \multirow{2}{2em}{10:1} & 
    0.5 & 0 & 95.24 & 0 & 0.5 & 0 & 95.24 & 0 \\
    & 1 & 0 & 95.24 & 0 & 1 & 0 & 95.24 & 0 \\
    \hline
    
    \multirow{2}{2em}{50:1} & 
    0.5 & 0 & 99.01 & 0 & 0.5 & 0 & 99.01 & 0 \\
    & 1 & 0 & 99.01 & 0 & 1 & 0 & 99.01 & 0 \\
    \hline
    
    \multirow{2}{2em}{100:1} & 
    0.5 & 0 & 99.50 & 0 & 0.5 & 0 & 99.50 & 0 \\
    & 1 & 0 & 99.50 & 0 & 1 & 0 & 99.50 & 0 \\
    \hline
    
    \multirow{2}{2em}{250:1} & 
    0.5 & 0 & 99.80 & 0 & 0.5 & 0 & 99.80 & 0 \\
    & 1 & 0 & 99.80 & 0 & 1 & 0 & 99.80 & 0 \\
    \hline
    
    \end{tabular}
\end{table}

In each cell, value of `0' 
implies that there is no correct prediction for that class. The above results show that both Broad Learning and Random Forest handle imbalanced data better than a Multilayer Perceptron. Table \ref{RF} shows that performance of Random Forest classifier is almost consistent across the various imbalance ratios, but its effectiveness is too much dependent on the choice of a suitable data-adjustment technique. In other words, Random Forest may not be able to perform well on any given data. On the other hand, although the effectiveness of Broad Learning depends on the choice of a suitable imbalance ratio, its performance is consistent for any given data adjustment technique, i.e., Broad Learning can robustly handle noisy data as long as the imbalance ratio is small. Similar to Random Forest, an ensemble of Broad Learners can be employed to study if the classification performance can be made robust to variation in imbalance ratio as well.

Working with Deep Neural Network requires a large amount of training data. 
The feature set for change detection, as described here, is of small dimension, and is not suitable for training a deep learning model.
Moreover, as shown in \cite{bls}, Broad Learning performs better and in a more efficient way on low-dimensional data as compared to that of Deep Learning Approaches. So, we have deliberately avoided comparison with Deep Learning in the present paper.

\section{Conclusions and Future Work}

In this paper, we describe a technique for detecting changes in aerial images over a period of time. Usual characteristics of a change detection ground truth data set is that, the number of pixels in changed category is significantly less than that in the unchanged category. Due to this fact, a classifier trained on the available ground truth data gets highly biased towards accurately classifying the majority class. We handle this problem by applying a suitable data adjustment technique on the imbalanced ground truth data to make it balanced. Experimental results show that the Broad Learning system works quite effectively 
in terms of the F-Score metric. 
Moreover, the results obtained from Random Under-sampling and SMOTE data adjustment technique is usually better than those obtained from a combination of random under-sampling and over-sampling technique for any classifier. 
In future, this technique can be also extended to change detection in videos that can be an efficient tool in surveillance. Good performance on the chosen data set also motivates us to test our approach on real satellite images. Broad Learning based change detection technique can also be employed in summarizing aerial videos.

\section*{Acknowledgements} The authors would like to thank Dr. Lipo Wang of Nanyang Technological University and Dr. C.L. Philip Chen of University of Macau for introducing us to the concept of Broad Learning and helping us with the initial implementation stage. Authors also sincerely acknowledge Dr. Prasenjit Banerjee, Nalanda Foundation Mentor for his support. Authors also share their sincere gratitude to HP Innovation Incubator program for giving them the opportunity towards initiation of this project.

\bibliographystyle{ieeetr}
\bibliography{main.bib}

\end{document}